\documentclass[letterpaper]{article} 
\usepackage{aaai2026}  
\usepackage{times}  
\usepackage{helvet}  
\usepackage{courier}  
\usepackage[hyphens]{url}  
\usepackage{graphicx} 
\usepackage{natbib}  
\usepackage{caption} 
\usepackage{algorithm}
\usepackage{algorithmic}

\usepackage{xcolor}
\usepackage{booktabs}
\usepackage{multirow}
\usepackage{bm, amsmath}
\usepackage{amssymb}

\def\equationautorefname{Eq.}%
\def\figureautorefname{Fig.}%
\def\tableautorefname{Table }%

\newcommand{\valstd}[2]{$#1 {\scriptstyle \,\pm\, #2}$}
\newcommand{\valstdb}[2]{$\mathbf{#1} {\scriptstyle \,\pm\, #2}$}
\newcommand{\valstdu}[2]{$\underline{#1} {\scriptstyle \,\pm\, #2}$}

\usepackage{newfloat}
\usepackage{listings}
\DeclareCaptionStyle{ruled}{labelfont=normalfont,labelsep=colon,strut=off} 
\floatstyle{ruled}
\newfloat{listing}{tb}{lst}{}
\floatname{listing}{Listing}
\title{Beyond Observations: Reconstruction Error-Guided Irregularly Sampled Time Series  Representation Learning}
\author{
    Jiexi Liu\textsuperscript{\rm 1}\equalcontrib,
    Meng Cao\textsuperscript{\rm 2,3}\equalcontrib,
    Songcan Chen\textsuperscript{\rm 2,3}\thanks{Corresponding Author}
}
\affiliations{
    \textsuperscript{\rm 1}School of Computer and Artificial Intelligence, Nanjing University of Finance and Economics\\
    \textsuperscript{\rm 2}College of Computer Science and Technology, Nanjing University of Aeronautics and Astronautics\\
    \textsuperscript{\rm 3}MIIT Key Laboratory of Pattern Analysis and Machine Intelligence\\   

    \{liujiexi, meng.cao, s.chen\}@nuaa.edu.cn%
}

\usepackage{bibentry}

\begin{document}

\maketitle

\begin{abstract}
Irregularly sampled time series (ISTS), characterized by non-uniform time intervals with natural missingness, are prevalent in real-world applications. Existing approaches for ISTS modeling primarily rely on observed values to impute unobserved ones or infer latent dynamics. However, these methods overlook a critical source of learning signal:\textit{ the reconstruction error inherently produced during model training.} Such error implicitly reflects how well a model captures the underlying data structure and can serve as an informative proxy for unobserved values. To exploit this insight, we propose \textbf{iTimER}, a simple yet effective self-supervised pre-training framework for ISTS representation learning. iTimER models the distribution of reconstruction errors over observed values and generates pseudo-observations for unobserved timestamps through a mixup strategy between sampled errors and the last available observations. This transforms unobserved timestamps into \textit{noise-aware training targets}, enabling meaningful reconstruction signals. A Wasserstein metric aligns reconstruction error distributions between observed and pseudo-observed regions, while a contrastive learning objective enhances the discriminability of learned representations. Extensive experiments on classification, interpolation, and forecasting tasks demonstrate that iTimER consistently outperforms state-of-the-art methods under the ISTS setting.
\end{abstract}

\section{Introduction}
\label{sec:introduction}
\begin{figure}[t]
\centerline{\includegraphics[width=0.9\linewidth]{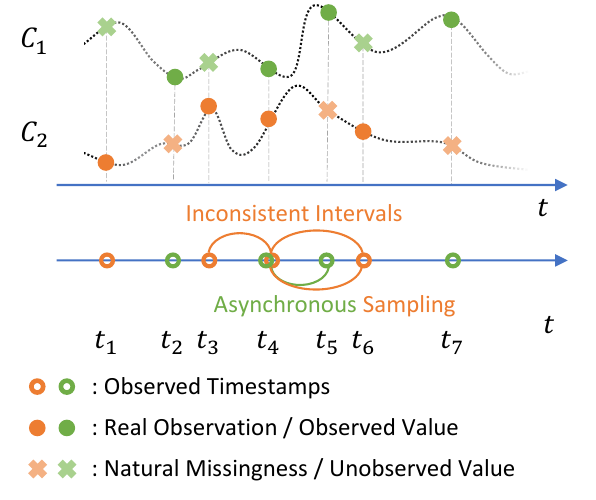}}
\caption{Multivariate irregularly sampled time series (ISTS) with two variables $C_1$ and $C_2$, exhibiting non-uniform sampling intervals with natural missingness. }
\label{fig:ISTS}
\end{figure}

Irregularly sampled time series (ISTS) are ubiquitous in real-world applications, ranging from healthcare \cite{goldberger2000physiobank,reyna2020early,liu2025timecheat}, meteorology \cite{schulz1997spectrum,cao2018brits} to transportation \cite{chen2022nonstationary, tang2020joint}. Arising from factors such as sensor failures, transmission distortions, or cost-driven acquisition policies, ISTS exhibit inconsistent intervals between consecutive sampling timestamps within a variable, asynchronous sampling across multiple variables and sometimes sampling sparsity. As shown in \figureautorefname \ref{fig:ISTS}, these characteristics manifest as \textbf{natural missingness}, i.e., unobserved values occurring, violating the assumption of a coherent, fixed-dimensional feature space that underlies most conventional time series analysis methods \cite{shukla2020survey}.

Recent studies have explored multiple directions to address the above challenge and model temporal dependencies with missing patterns. A common line of work leverages explicit imputation of missing values as a preprocessing step before learning representations or performing downstream tasks~\cite{che2018recurrent, camino2019improving, tashiro2021csdi, zhang2021life, chen2022nonstationary, fan2022dynamic, du2023saits, fons2025lscd}. However,  such direct imputation can be unreliable under high missingness ratio and may introduce noise or bias~\cite{zhang2021graph, wu2021dynamic, agarwal2023modelling, sun2024time}. Therefore, other kind of method proposes end-to-end architectures that typically model the data using continuous-time dynamics~\cite{kidger2020neural, rubanova2019latent, jhin2022exit, oh2025comprehensive, zhang2025diffode} or learning discrete regularly-sampled representations~\cite{shukla2021multitime,zhang2023warpformer,yalavarthi2024grafiti,agarwal2023modelling,zhang2024tpatchgnn}. In parallel, a growing body of work adopts self-supervised learning paradigms for ISTS, where supervision is derived from the data itself ~\cite{chowdhury2023primenet,beebe2023paits}. These methods apply contrastive learning to align representations of different augmented views, while others design reconstruction-based objectives from perturbed dropout inputs to encourage informative representations.

Despite the above advancements, existing approaches for learning from ISTS exhibit a fundamental limitation: they predominantly rely on the observed data as the only available source of training signals, except for label information. As a result, the model’s behavior in unobserved regions remains underconstrained, and its representational potential is largely overlooked. Crucially, these approaches tend to disregard model-intrinsic signals that arise naturally during optimization. Among them, the reconstruction error that quantifies the discrepancy between inputs and outputs encodes implicit information about the model’s current understanding of the data. While self-supervised masked modeling methods usually minimize such error over observed regions, they typically treat it as a loss term, rather than a source of feedback that can inform learning beyond the observations. In our opinion, the reconstruction error itself carries rich statistical cues that can be leveraged to guide learning in unobserved regions. In particular, the error distribution reflects the model’s uncertainty and inductive bias, offering a unique opportunity to synthesize auxiliary training signals without relying on external supervision information or heuristic augmentations. Motivated by this insight, in this work, we propose \textbf{iTimER}, a novel self-supervised pretraining framework that treats reconstruction error as an informative learning signal. By modeling its distribution and propagating it to unobserved timestamps, we construct pseudo-observations that enable distributionally aligned learning. This strategy not only enriches the training signal but also paves the way for more reliable and expressive representations in ISTS without requiring any labeled data.

Rather than merely imputing missingness or injecting arbitrary noise,  iTimER leverages the model’s own reconstruction behavior as a proxy for learning uncertainty. Specifically, we estimate a distribution of reconstruction errors over the observed timestamps and propagate this signal to unobserved regions via a sampling-and-mixup mechanism. This strategy reflects both the model’s uncertainty and its inductive preferences, enabling the generation of pseudo-observations aligned with the model's learned reconstruction error distribution. As observations in ISTS are not uniformly distributed over time but are often concentrated around specific events or high-frequency periods, leaving other segments completely unobserved. As a result, the observed data only provide a partial view of the series, leading models trained on these inputs to learn biased dynamics that can impair generalization and downstream performance. Although some recent methods~\cite{tashiro2021csdi,islam2025self} attempt to mitigate this issue by adding and removing noise during training, they typically overlook the uneven, biased nature of real-world sampling. In contrast, iTimER constructs pseudo-observations based on the model’s own uncertainty, enabling a noise-aware learning signal. Rather than treating the observed data as perfect learning signals, iTimER embraces its imperfections and uses them as a proxy to guide representation learning for ISTS.

To further ensure consistency between real and pseudo signals, iTimER imposes a Wasserstein metric constraint between the reconstruction error distributions of observed and pseudo-observed regions. This encourages the model to maintain similar uncertainty patterns across both parts of the sequence, effectively mitigating the impact of sampling bias. Additionally, iTimER employs contrastive learning to enhance the discriminative power of the learned representations, enabling the encoder to better capture meaningful variations in temporal dynamics.

To sum up, iTimER uses self-generated, noise-aware supervision signals that reveal how the model generalizes beyond the observed data to learn meaningful representations under ISTS conditions. Our main contributions are summarized as follows:

\begin{itemize}
    \item We uncover the untapped value of reconstruction error as an informative self-supervised signal, providing a novel perspective for representation learning in ISTS.
    \item We enforce consistency between real and pseudo-observation series by aligning their reconstruction error distributions, improving the reliability for better representation learning.
    \item We present a task-agnostic pre-training model for ISTS, applicable to various downstream tasks, including classification, forecasting, and interpolation.
\end{itemize}

\section{Related Work}

\paragraph{Representation Learning for Irregularly Sampled Time Series}

Recent advances in ISTS analysis have increasingly shifted toward learning expressive and discriminative representations. A common kind of imputation-based methods impute unobserved values from real observations before downstream tasks~\cite{che2018recurrent, chen2022nonstationary, fan2022dynamic, du2023saits}.  However, inappropriate imputation can introduce misleading noise or structural bias, especially under sparse observations that negatively impact model performance~\cite{zhang2021graph, agarwal2023modelling}. While end-to-end models aggregate observation values for each variable to get discrete hidden states using attention mechanisms~\cite{shukla2021multitime, zhang2023warpformer, zhou2025revitalizing}, Graph Neural Networks (GNNs)~\cite{zhang2021graph, yalavarthi2024grafiti, zhang2024tpatchgnn}, and Recurrent Neural Networks (RNNs)~\cite{de2019gru, schirmer2022modeling, agarwal2023modelling} -based methods. Such methods struggle to capture continuous temporal dependencies. To address this, Neural ODE-based approaches~\cite{kidger2020neural, rubanova2019latent, jhin2022exit, oh2025comprehensive, zhang2025diffode} have been proposed to model ISTS in continuous time.

Recent work has explored self-supervised learning for ISTS representation learning by deriving supervision directly from the data itself~\cite{chowdhury2023primenet, beebe2023paits}. Self-supervised methods design pretext tasks that encourage the model to learn meaningful representations without relying on downstream labels, including contrastive learning that constructs augmented views from real observations for contrastive learning and masked modeling methods that drop parts of the input data, and predict the dropout values by minimizing the reconstruction error between the predicted and original values. When facing ISTS, the non-uniform time intervals with natural missingness make it difficult to construct meaningful positive pairs and waste valuable real observations and potentially distort the learning signal.

Different from existing masked modeling methods, iTimER leverages reconstruction error not just as a loss, but as a learning signal to model uncertainty. It estimates error distributions from observed data and synthesizes pseudo-observations, enabling noise-aware representation learning without explicit imputation or artificial noise~\cite{tashiro2021csdi,islam2025self}.

\subsection{Reconstruction Error as a Learning Signal}

Across domains, reconstruction error, also referred to as residual, has proven to be more than a loss metric but serves as a valuable signal for model guidance. In image inpainting, \citet{yu2018generative} employs reconstruction error in a coarse-to-fine architecture, where the residuals from the coarse stage are passed to the fine stage as structural cues, improving detail consistency in the final output. Similarly, in robust subspace learning, \citet{meng2013robust} embeds reconstruction error into the optimization loop to jointly learn the data structure and noise characteristics. In video anomaly detection, works such as \citet{hasan2016learning} and \citet{liu2018future} exploit spatially localized residuals to identify abnormal events.

These examples suggest that reconstruction error can highlight uncertainty, noise, or outliers and serve as an informative learning cue. iTimER builds on this idea by modeling the distribution of reconstruction error over observed regions in ISTS, then sampling from this distribution to generate uncertainty-aware pseudo-observations. These are used not as direct labels, but as a proxy signal that drives both contrastive representation learning and distributional consistency. To our knowledge, iTimER is the first to formalize this strategy in the ISTS domain, using reconstruction error not as a penalty, but as a means of self-supervised guidance.

\begin{figure*}[t]
\centerline{\includegraphics[width=0.9\linewidth]{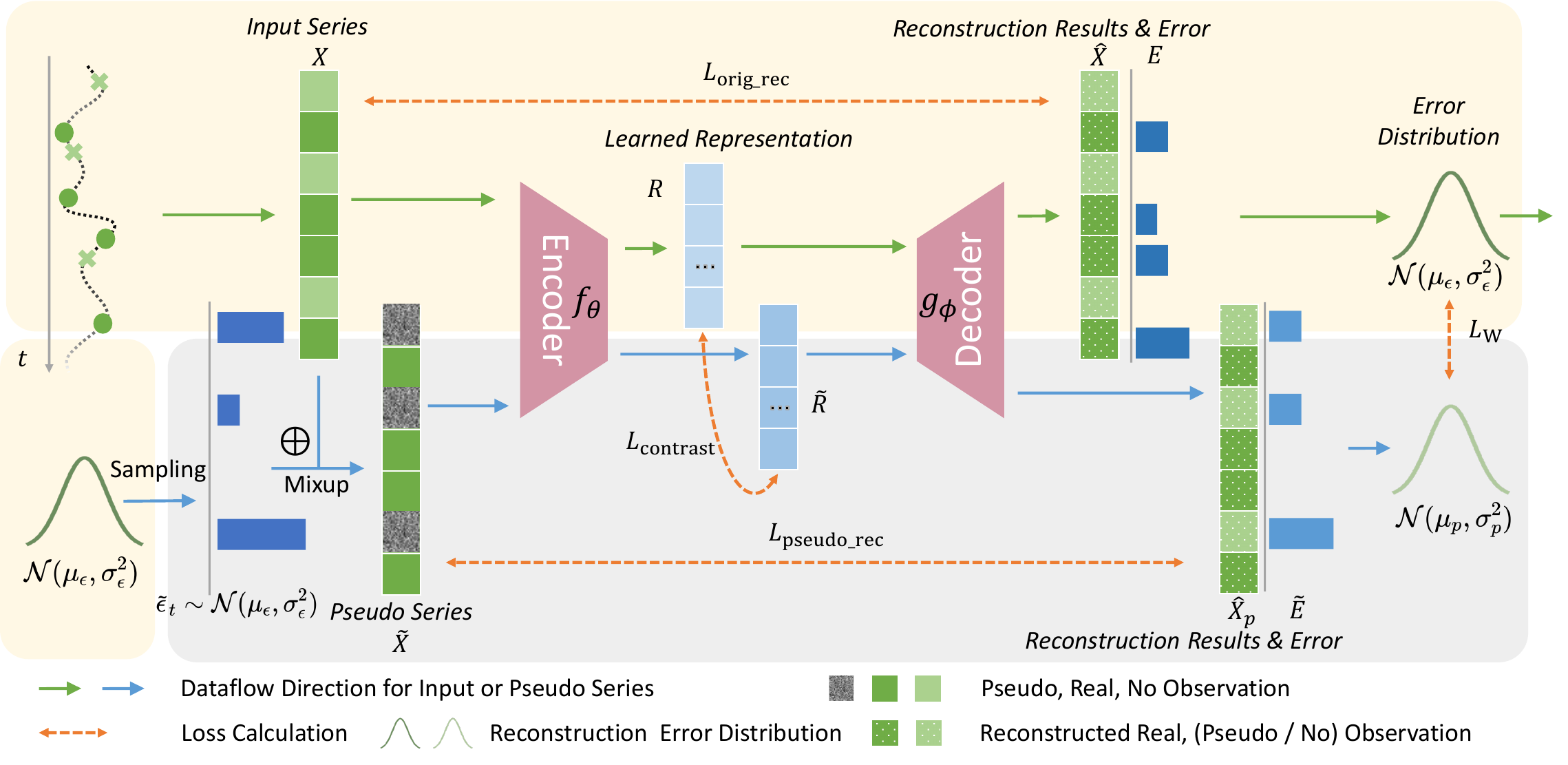}}
\caption{\textbf{The iTimER framework. }iTimER leverages reconstruction error from the original sequence to generate pseudo observations via mixup uncertainty-aware sampling and the last observed value. Both real and pseudo-observation series are encoded and reconstructed, with consistency enforced at both representation and error distribution levels.}
\label{fig:framework}
\end{figure*}

\section{Proposed iTimER Framework}

 \subsection{Problem Setup and Method Outline} 
Let $\mathcal{X} = \{ \bm{X}_1, \bm{X}_2, \dots, \bm{X}_N \}$ denote a dataset of $N$ ISTS instances, where each instance $\bm{X}_n \in \mathbb{R}^{T \times C}$ contains $C$ variables observed over $T$ time steps. The binary mask $\bm{M}_n \in \{0,1\}^{T \times C}$ is a missingness indicator for $\bm{X}_n$, where $m_{n,t,c} = 0$ indicates there is no observation at timestamp $t$ in variable $c$ and $m_{n,t,c} = 1$ means it is observed. Our objective is to learn a nonlinear encoding function $f_\theta: \mathbb{R}^{T \times C} \to \mathbb{R}^{\tau \times D}$ that maps each ISTS instance into a representation $\bm{R}_n = f_\theta(\bm{X}_n)$, where $\tau$ is the length of the representation and $D$ is the dimension. \textit{ Here, for brevity, we omit the data case index $n$ for the $n$-th instance and $c$ for the $c$-th variable in an instance when the context is clear. }

Specifically, as illustrated in \figureautorefname{\ref{fig:framework}}, given an ISTS instance $\bm{X}$ with its missingness indicator $\bm{M}$, we employ an encoder-decoder architecture, where $f_\theta$ maps the input series to a latent representation $\bm{R}$ and $g_\phi$ reconstructs the input from it to obtain reconstruction $\widehat X$. We first estimate the reconstruction error distribution $\epsilon_{t} \sim \mathcal{N}(\mu_\epsilon, \sigma_\epsilon^2)$ from $x_t-\hat x_t$, where $m_{t}=1$. Then, for unobserved timestamps, i.e., $m_{t}=0$, we sample errors $\tilde\epsilon_{t}$ from $\mathcal{N}(\mu_\epsilon, \sigma^2_\epsilon)$ and mixup it with the last available observation $x_{t-1}$ to obtain the pseudo-observation $\tilde{x}_t$. This results finally becoming the pseudo-observation series $\widetilde{\bm{X}}$ that preserves the original timestamps but contains pseudo values in place of the unobserved timestamps. 

By jointly optimizing distribution alignment, representation consistency and reconstruction, this training strategy guides the encoder toward learning more informative and discriminative representations $\bm{R}$ for downstream tasks, and we will elaborate on the key steps in the following sections.

\subsection{Pseudo-Observation Learning}
\paragraph{Modeling Reconstruction Uncertainty.}
For any instance $\bm{X}$, to model the uncertainty of reconstruction, we begin by reconstructing the observed timestamps, i.e., mask $m_t=1$ by using an encoder-decoder model $\hat{x}_t = g_\phi(f_\theta(x_t))$. The reconstruction error is then computed as $\epsilon_t = x_t - \hat{x}_t$. To capture its statistical properties, we assume the error distribution $\epsilon_t$ following a Gaussian distribution $\epsilon_t\sim\mathcal{N}(\mu_\epsilon, \sigma_\epsilon)$, which can provide a basis for generating plausible pseudo-observations for the unobserved timestamps.

We leverage a Gaussian form for the reconstruction error distribution, which aligns with the empirical observation that residuals from deep time series models tend to follow near-normal patterns. This is also suitable for ISTS \cite{shukla2021multitime,fortuin2020gp}, where the stochastic nature of sampling times introduces non-uniform uncertainty. A Gaussian approximation allows us to quantify such uncertainty and synthesize pseudo-observations in the absence of complete data.

In addition, to preserve informative signals while maintaining stability, a momentum strategy is adopted to utilize historical information through a momentum coefficient $\rho$ for the reconstruction error as
\begin{equation}
    \mu_\epsilon^h = \rho\cdot \mu_\epsilon^{h-1} + (1-\rho) \cdot \mu_\epsilon, \quad \sigma_\epsilon^h = \rho\cdot \sigma_\epsilon^{h-1} + (1-\rho) \cdot \sigma_\epsilon, 
     \label{eq:momentum}
\end{equation}
where $\mu_\epsilon^{h-1}$ and $\sigma_\epsilon^{h-1}$ denote the mean and standard deviation at the $(h-1)$-th iteration, separately.

\paragraph{Error Sampling at Unobserved Timestamps.}
For unobserved timestamps, i.e., $m_t = 0$, we synthesize pseudo-observations by mixing a sampled reconstruction error with a temporal contextual anchor value, such as the global mean, local moving average, or the last observation value. In this paper, we choose the last observation as it can better preserve temporal continuity and aligns with the causal nature of time series. Compared to the global mean or local moving average, it is not only a simple processing, but also faithfully maintains local dynamics. Then, we obtain the Mixup strategy defined as
\begin{equation}
    \begin{aligned}
        \tilde{x}_t = \alpha_t \cdot \bar{x} + (1 - \alpha_t) \cdot \tilde \epsilon_t,\quad \tilde\epsilon_t\sim\mathcal{N}(\mu_\epsilon^h, (\sigma_\epsilon^h)^2)
     \end{aligned}
     \label{eq:mixup}
\end{equation}
where $\alpha_t \in [0, 1]$ controls the mixing ratio.  This approach generates softly regularized imputations that balance between global structure and instance-specific uncertainty. In ISTS, where temporal gaps vary significantly and structural assumptions may not hold, such a mixup-based generation helps avoid overconfident or abrupt imputations, encourages smoother continuity, and enhances representation learning by exposing the model to more realistic distributional variations.

\paragraph{Constructing Complete Pseudo-Observation Series.}
To form a complete pseudo-observation series, we integrate the sampled values with the original observations based on the observed mask. Specifically, for each time step $t$, if the original value $x_t$ is observed (i.e., $m_t = 1$), we retain the original observation. If $x_t$ is missing (i.e., $m_t = 0$), we replace it with the generated pseudo-observation $\tilde{x}_t$ introduced above. Formally, the constructed series is defined as:
\begin{equation}
  \tilde{x}_t = \begin{cases} 
    x_t, & \text{if } m_t = 1 \\
    \tilde{x}_t, & \text{if } m_t = 0 
  \end{cases}
     \label{eq:pseudo}
\end{equation}

This procedure ensures that the final sequence preserves all original valid observations while providing realistic, distribution-aware proxy for the missingness, enabling the model to learn from a more complete and semantically coherent input. $\widetilde{\bm{X}}$ then follows the similar procedure that reconstructed by $g_
\phi(f_\theta(\widetilde{\bm{X}}))$ and obtain the reconstruction error distribution $ \mathcal{N}(\mu_p, \sigma^2_p)$ where $m_t=0$.

\subsection{Learning Objective Design }
Our main step of iTimER is to enforce reconstruction error consistency. The unobserved timestamps in ISTS have no ground truth and exhibit non‐uniform sampling intervals that cannot directly supervise the model learning in those regions. By matching the statistical profile of errors between observed and imputed points, we provide a principled learning signal on the unobserved timestamps that 1) quantifies the model’s uncertainty in unobserved timestamps, 2) prevents overconfident, biased imputations, and 3) encourages the encoder to learn representations that generalize across both observed and missing parts. This error distribution alignment thus serves as a proxy for true reconstruction quality where observations are unavailable, leading to uncertainty‐aware ISTS representations. 

Here, to match the reconstruction error distributions of the real and pseudo-observation series, we use the 2-Wasserstein distance~\cite{panaretos2019statistical} between $ P_r=\mathcal{N}(\mu_{\epsilon}, \sigma^2_{\epsilon})$ and $ P_p = \mathcal{N}(\mu_p, \sigma^2_p)$ as follows
\begin{equation}
\begin{aligned}
  L_{\text{W}}  & = D_{\text{2-Wasserstein}}(P_r \parallel P_p)\\
  & = \left\|\mu_{\epsilon}-\mu_{p}\right\|^{2}+\left\|\sigma_{\epsilon}-\sigma_{p}\right\|^{2}
\end{aligned}
     \label{eq:lw}
\end{equation}

Moreover, to ensure robust and informative representation learning under ISTS, we incorporate both contrastive learning and dual reconstruction objectives. The contrastive loss encourages the latent representations of the original and pseudo-observation series to remain close, thereby mitigating representation drift caused by missing values. This alignment in the embedding space not only enhances the semantic consistency between observed and reconstructed data but also facilitates the learning of stable, discriminative features that are less sensitive to sampling irregularities or noise.

Input ISTS $\bm{X}$ and pseudo-observation series $\widetilde{\bm{X}}$ are encoded using a shared encoder $f_\theta(\cdot)$, producing latent representations $\bm{R}$ and $\widetilde{\bm{R}}$, respectively:
\begin{equation}
   \bm{R} = f_\theta(\bm{X}), \quad \widetilde{\bm{R}} = f_\theta(\widetilde{\bm{X}}) 
     \label{eq:encode}
\end{equation}
Therefore, the contrastive loss function is:
\begin{equation}
\small
    L_{\text{contrast}}=-\sum\nolimits_{i=1}^{|\mathcal{B}|}\log \frac{\exp \left(\boldsymbol{R}_i \cdot {\widetilde{\bm{R}}_i}\right)}{\sum_{j=1}^{|\mathcal{B}|}\left((\boldsymbol{R}_i \cdot \widetilde{\bm{R}}_i)+\mathbb{I}_{[i \neq j]} \left(\boldsymbol{R}_i \cdot {\bm{R}_j}\right)\right)}
    \label{con}
\end{equation}
where the $\mathbb{I}$ is the indicator function, $|\mathcal{B}|$ indicates batch size.  

The dual reconstruction losses \equationautorefname \eqref{eq:reco}, one on the original observations and the other on the pseudo-observation series, serve complementary roles. The original reconstruction loss serves as a fundamental self-supervision signal to ensure that the encoder $f_\theta$ and decoder $g_\phi$ can effectively capture the underlying structure of the observed data and accurately reconstruct the ISTS. While the pseudo-observation reconstruction loss evaluates whether the model can still produce stable and reasonable reconstructions. It strengthens the model’s ability to handle pseudo-observations, helping it learn the underlying distribution of the imputed regions.
\begin{equation}
    \begin{aligned}
        &  L_{\text{orig\_rec}} = \|\bm{M} \odot (\bm{X} - \widehat{\bm{X}}) \|^2 \\
        &   L_{\text{pseudo\_rec}} = \| \bm{M} \odot  ( \widetilde{\bm{X}} - \widehat{\bm{X}}_{\text{p}}) \|^2
    \end{aligned}
    \label{eq:reco}
\end{equation}

 The overall loss function is a weighted combination of the above terms:
\begin{equation}
    \begin{aligned}
L = \alpha L_{\text{W}} + \beta L_{\text{contrast}} + \frac{1}{2} (L_{\text{orig\_rec}} +  L_{\text{pseudo\_rec}}) 
    \end{aligned}
\end{equation} 
where $\alpha$ and $\beta$ are hyperparameters controlling the relative importance of each term.

\begin{table*}[!t]
\small
\centering

\begin{tabular}{l|cc|cc|cccc}
\toprule
& \multicolumn{2}{c|}{P12} & \multicolumn{2}{c|}{P19} & \multicolumn{4}{c}{PAM} \\ \cmidrule{2-9}
\multirow{-2}{*}{Method} & AUROC & AUPRC & AUROC & AUPRC & Accuracy & Precision & Recall & F1 score \\ \midrule
Transformer & \valstd{83.3}{0.7} & \valstd{47.9}{3.6} & \valstd{80.7}{3.8} & \valstd{42.7}{7.7} & \valstd{83.5}{1.5} & \valstd{84.8}{1.5} & \valstd{86.0}{1.2} & \valstd{85.0}{1.3} \\
MTGNN & \valstd{74.4}{6.7} & \valstd{35.5}{6.0} & \valstd{81.9}{6.2}  & \valstd{39.9}{8.9} & \valstd{83.4}{1.9} & \valstd{85.2}{1.7} & \valstd{86.1}{1.9} & \valstd{85.9}{2.4}\\
DGM$^2$-O & \valstd{84.4}{1.6} & \valstd{47.3}{3.6} & \valstd{86.7}{3.4} & \valstd{44.7}{11.7} & \valstd{82.4}{2.3} & \valstd{85.2}{1.2} & \valstd{83.9}{2.3} & \valstd{84.3}{1.8}\\
IP-Net& \valstd{82.6}{1.4} &\valstd{47.6}{3.1} & \valstd{84.6}{1.3} & \valstd{38.1}{3.7} &\valstd{74.3}{3.8} & \valstd{75.6}{2.1} & \valstd{77.9}{2.2} & \valstd{76.6}{2.8}\\
GRU-D & \valstd{81.9}{2.1} & \valstd{46.1}{4.7} & \valstd{83.9}{1.7} & \valstd{46.9}{2.1} & \valstd{83.3}{1.6} & \valstd{84.6}{1.2} & \valstd{85.2}{1.6} & \valstd{84.8}{1.2}\\
SeFT & \valstd{73.9}{2.5} & \valstd{31.1}{4.1} & \valstd{81.2}{2.3} & \valstd{41.9}{3.1} & \valstd{67.1}{2.2} & \valstd{70.0}{2.4} & \valstd{68.2}{1.5} & \valstd{68.5}{1.8}\\
mTAND & \valstd{84.2}{0.8} & \valstd{48.2}{3.4} & \valstd{84.4}{1.3} & \valstd{50.6}{2.0} & \valstd{92.9}{0.8} & \valstd{93.8}{0.8} & \valstd{94.0}{0.9} & \valstd{93.8}{0.8}\\ 
Raindrop & \valstd{82.8}{1.7} & \valstd{44.0}{3.0} & \valstd{87.0}{2.3} & \valstd{51.8}{5.5} &\valstd{88.5}{1.5} & \valstd{89.9}{1.5} & \valstd{89.9}{0.6} & \valstd{89.8}{1.0}\\
Warpformer &\valstd{83.4}{0.9}  &\valstd{47.2}{3.7} &\valstdu{88.8}{1.7} &\valstdb{55.2}{3.9} & \valstd{94.3}{0.6} &\valstd{95.8}{0.8} &\valstd{94.8}{1.0} & \valstd{95.2}{0.6}\\
ViTST &\valstdu{85.1}{0.8} &\valstdu{51.1}{4.1} &\valstdb{89.2}{2.0} &\valstdu{53.1}{3.4} & \valstdu{95.8}{1.3} &\valstdu{96.2}{1.3} &\valstdu{96.1}{1.1} & \valstdu{96.5}{1.2} \\
\cmidrule{1-9}
FPT &\valstd{84.8}{1.1}  &\valstd{50.7}{3.0} &\valstd{87.3 }{2.9} &\valstd{51.6}{3.6} & \valstd{94.0}{1.4} &\valstd{95.3}{0.9} &\valstd{94.7}{1.1} & \valstd{94.9}{1.1}\\
Time-LLM &\valstd{84.4}{1.8} &\valstd{50.2}{1.6} &\valstd{85.1}{2.6} &\valstd{50.1}{3.4} &\valstd{93.4}{1.2} &\valstd{94.2}{1.3} &\valstd{94.7}{1.0} &\valstd{94.4}{1.1} \\


PrimeNet & \valstd{84.9}{0.6} & \valstd{49.8}{2.7} & \valstd{84.4}{1.3} & \valstd{39.7}{3.1} & \valstd{95.3}{0.5} & \valstd{96.1}{0.3} & \valstd{95.5}{0.6} & \valstd{95.7}{0.4} \\

\cmidrule{1-9}
\textbf{iTimER} &\valstdb{85.7}{0.8}  &\valstdb{52.0}{2.1} &\valstd{87.1}{0.6} &\valstd{45.6}{3.5} & \valstdb{96.1}{0.8} &\valstdb{96.7}{0.5} &\valstdb{96.4}{0.9} & \valstdb{96.6}{0.7}\\
\bottomrule
\end{tabular}
\caption{Overall performance comparison on ISTS \emph{Classification} task.}
\label{tab:main_result_classify} 

\end{table*}

\begin{table*}[!t]
    \centering
    \begin{minipage}[t]{0.475\textwidth}
        \centering
        \tabcolsep 0.45em
        \begin{tabular}{l|c|c|c}
            \toprule
            &PhysioNet & MIMIC& Human Activity \\
            \cmidrule{2-4} 
            \multirow{-2}{*}{Method} & MSE$\times10^{-3}$  & MSE$\times10^{-2}$  & MSE$\times10^{-3}$ \\
            \midrule
            GRU-D &\valstd{6.18}{0.23}  &\valstd{2.06 }{0.05 }     &\valstd{2.74}{0.09}  \\
            SeFT &\valstd{9.46}{0.12}  &\valstd{2.12 }{0.02 }  &\valstd{14.95}{0.03}  \\
            Raindrop &\valstd{10.65}{0.12}  &\valstd{2.31 }{0.04 }  &\valstd{15.21}{0.12}  \\
            Warpformer &\valstd{6.37}{0.34}  &\valstd{1.93 }{0.06 }  &\valstd{2.59}{0.15}  \\
            mTAND &\valstd{5.65}{0.08}  &\valstd{1.93 }{0.05 }  &\valstd{2.07}{0.17}  \\
            Latent-ODE &\valstd{6.84}{0.34}  &\valstd{1.89 }{0.08 }  &\valstd{3.12}{0.22}  \\
            Neural Flow &\valstd{6.77 }{0.06 }  &\valstd{2.18 }{0.11 }  &\valstd{3.73}{0.06}  \\
            CRU &\valstd{10.30}{0.10}  &\valstd{2.52 }{0.04 }  &\valstd{7.17}{0.32}  \\
            t-PatchGNN &\valstd{4.75}{0.03}   &\valstd{1.55 }{0.08 }  &\valstd{1.95 }{0.12 } \\
            \cmidrule{1-4}
            FPT &\valstd{12.24}{0.07}  &\valstd{3.71 }{0.01 }  &\valstd{2.85}{0.09}  \\
            Time-LLM &\valstd{12.43 }{0.08 }  &\valstd{3.63 }{0.05 }  &\valstd{2.92 }{0.01 }  \\
            ISTS-PLM &\valstdu{4.55}{0.08}  &\valstdu{1.47 }{0.01}  &\valstdu{1.93}{0.01}  \\
            \cmidrule{1-4}
            \textbf{iTimER} & \valstdb{2.86}{0.04}  & \valstdb{0.13}{0.00}  & \valstdb{1.82}{0.01} \\
            \bottomrule
        \end{tabular}
        \caption{Overall performance comparison on ISTS \emph{Interpolation} task.}
        \label{tab:main_result_interpolation}
    \end{minipage}
    \hspace{1.5em}
    \begin{minipage}[t]{0.475\textwidth}
        \centering
        \tabcolsep 0.45em
        \begin{tabular}{l|c|c|c}
            \toprule
            &PhysioNet & MIMIC& Human Activity \\
            \cmidrule{2-4} 
            \multirow{-2}{*}{Method} & MSE$\times10^{-3}$  & MSE$\times10^{-2}$  & MSE$\times10^{-3}$ \\
            \midrule
            GRU-D & \valstd{5.59}{0.09}  &\valstd{1.76}{0.03}  &\valstd{2.94}{0.05} \\
            SeFT & \valstd{9.22}{0.18}  &\valstd{1.87}{0.01}  &\valstd{12.20}{0.17} \\
            Raindrop & \valstd{9.82}{0.08}  &\valstd{1.99}{0.03}  &\valstd{14.92}{0.14} \\
            Warpformer &\valstd{5.94}{0.35}  &\valstd{1.73}{0.04}  &\valstd{2.79}{0.04}  \\
            mTAND &\valstd{6.23}{0.24}   &\valstd{1.85}{0.06}  &\valstd{3.22}{0.07} \\
            Latent-ODE &\valstd{6.05}{0.57}  &\valstd{1.89}{0.19}  &\valstd{3.34}{0.11} \\
            Neural Flow &\valstd{7.20}{0.07}  &\valstd{1.87}{0.05}  &\valstd{4.05}{0.13}  \\
            CRU &\valstd{8.56}{0.26}   &\valstd{1.97}{0.02}  &\valstd{6.97}{0.78} \\
            t-PatchGNN &\valstd{4.98}{0.08}   &\valstd{1.69}{0.03}  &\valstd{2.66}{0.03} \\
            \cmidrule{1-4}
            FPT &\valstd{10.95}{0.02}   &\valstd{4.00 }{0.03 }  &\valstd{3.03}{0.09} \\
            Time-LLM &\valstd{11.56 }{0.19 }   &\valstd{4.41}{0.01 } &\valstd{3.21 }{0.01} \\
            ISTS-PLM &\valstd{4.92 }{0.05 }  &\valstd{1.64}{0.02 }  &\valstdb{2.58}{0.03} \\
            \cmidrule{1-4}
            \textbf{iTimER} & \valstdb{3.64}{0.05}  & \valstdb{0.14}{0.00}  & \valstdu{2.75}{0.03}  \\
            \bottomrule
        \end{tabular}
        \caption{Overall performance comparison on ISTS \emph{Forecasting} task.}
        \label{tab:main_result_predict}
    \end{minipage}
\end{table*}

\section{Experiments}
 \label{experiment}
In this section, we show the effectiveness of the iTimER framework across $3$ mainstream time series downstream tasks, including classification, interpolation, and forecasting. The results are reported as mean and standard deviation values calculated over $5$ independent runs. We use Time-Feature Attention (TFA) and Feature-Feature Attention (FFA) as the backbone encoder \cite{chowdhury2023primenet}. The \textbf{bold} font highlights the top-performing method, while the \underline{underlined} text marks the runner-up. Additional experimental setup and encoder details are provided in the Appendix due to space constraints.

\subsection{Experimental Setup}

\subsubsection{Datasets and Metrics.}
For the \emph{Classification} task,
we evaluate on three real-world ISTS datasets from healthcare and human activity domains: 
(1) \textbf{P19} \cite{reyna2020early} includes $38,803$ ICU patients monitored by $34$ sensors, with a missing ratio of $94.9\%$.
(2) \textbf{P12} \cite{goldberger2000physiobank} contains temporal measurements from $11,988$ patients over the first $48$ hours of ICU stay, using $36$ sensors and a missing ratio of $88.4\%$.
(3) \textbf{PAM} \cite{reiss2012introducing} consists of $5,333$ sequences of $8$ daily activities measured by $17$ sensors, with $60.0\%$ missingness.

\textit{P19 and P12 are \textbf{imbalanced} binary label datasets} while PAM dataset is balance containing $8$ classes. Following the standard practice and prior works~\cite{li2023time, zhang2021graph}, we split each dataset into training/validation/test sets using an $8:1:1$ ratio and fixed indices for fair comparison. For the \textit{imbalanced} P12 and P19 datasets, we report Area Under the Receiver Operating Characteristic Curve (AUROC) and Area Under the Precision-Recall Curve (AUPRC), while for the nearly balanced PAM dataset, we use Accuracy, Precision, Recall, and F1 Score. Higher values across all these metrics indicate better performance.

For \emph{Interpolation and Forecasting} tasks,
we use three natural ISTS datasets from the healthcare and activity domains:
(1) \textbf{Physionet} \cite{goldberger2000physiobank} includes $12,000$ ICU patients monitored by $41$ sensors, with a missing ratio of $85.7\%$.
(2) \textbf{MIMIC} \cite{johnson2016mimic} contains $23,457$ ISTS samples covering the first $48$ hours of ICU stays, with $96$ variables and $96.7\%$ missingness.
(3) \textbf{Human Activity }\cite{frank2010uci}
comprises $5,400$ samples of 3D positional data collected from $12$ sensors, with $75.0\%$ missingness.

We randomly split each dataset into training, validation, and test sets with a $6:2:2$ ratio, applying min-max normalization to the original observation values. For both tasks, we evaluate performance using Mean Squared Error (MSE) and Mean Absolute Error (MAE), where lower values indicate better performance. Complete results are reported in the appendix.

\subsubsection{Baselines.}
To evaluate the performance of iTimER in ISTS 
\emph{Classification}, we incorporate the following baseline models for a fair comparison, including Transformer \cite{vaswani2017attention}, MTGNN \cite{wu2020connecting}, DGM$^2$-O \cite{wu2021dynamic}, IP-Net \cite{shukla2018interpolation}, GRU-D \cite{che2018recurrent}, SeFT \cite{horn2020set}, mTAND \cite{shukla2021multitime}, Raindrop \cite{zhang2021graph}, Warpformer \cite{zhang2023warpformer} and ViTST \cite{li2023time}. Specially, we also compare with several Pre-trained Language Model (PLM)-based methods for regularly sampled time series analysis, such as FPT \cite{zhou2023one} and Time-LLM \cite{jin2024timellm}, as well as pre-training ISTS models PrimeNet \cite{chowdhury2023primenet}.

For ISTS \emph{Interpolation and Forecasting} tasks, except adapting the representative baselines above to these two tasks, we further incorporate several models designed for the ISTS prediction tasks, including Latent-ODE \cite{rubanova2019latent}, Neural Flow \cite{bilovs2021neural}, CRU \cite{schirmer2022modeling}, t-PatchGNN \cite{zhang2024tpatchgnn} and ISTS-PLM \cite{zhang2024unleash}, a PLM-based method for ISTS.

\begin{table}[h]
    \centering
    \tabcolsep 0.45em
    \begin{tabular}{c|c|c|c}
        \toprule
        &P12 & PhysioNet-I& PhysioNet-F \\
        \cmidrule{2-4} 
        \multirow{-2}{*}{Method} & AUROC $\uparrow $ & MSE$\times10^{-3}$ $\downarrow $ & MSE$\times10^{-3}$ $\downarrow $ \\
        \midrule
        Baseline & \valstd{84.6}{2.1}  & \valstd{5.03}{0.04}  & \valstd{6.10}{0.29}  \\
        \midrule
        Random & \valstd{84.8}{0.6}  & \valstd{4.01}{0.06}  & \valstd{4.61}{0.10}  \\
        Constant & \valstd{84.7}{0.9}  & \valstd{3.89}{0.10}  & \valstd{4.26}{0.07}  \\
        Only Error & \valstd{85.1}{0.8}  & \valstd{3.08}{0.06}  & \valstd{3.77}{0.05}  \\
        \midrule
        Zero   & \valstd{85.0}{1.9}  & \valstd{3.06}{0.10}  & \valstd{3.94}{0.12}  \\
        Mean   & \valstd{85.4}{1.0} & \valstd{2.91}{0.06}  & \valstd{3.66}{0.08} \\
        MAve & \valstd{85.8}{1.0}  & \valstd{2.83}{0.01}  & \valstd{3.64}{0.02}  \\
        \midrule
        w/o $L_\text{W}$ & \valstd{85.0}{1.0}  & \valstd{3.17}{0.05}  & \valstd{3.91}{0.01}  \\
        w/o $L_\text{contrast}$ & \valstd{85.4}{0.7}  & \valstd{3.02}{0.05}  & \valstd{3.78}{0.03}  \\
        \midrule
        \textbf{iTimER} & \valstd{85.7}{0.8}  & \valstd{2.86}{0.04}  & \valstd{3.64}{0.05}  \\
        \bottomrule
    \end{tabular}
    \caption{Ablation results of iTimER on three downstream tasks, evaluating different pseudo-observation strategies and loss components (P12 for classification, PhysioNet-I for interpolation, and PhysioNet-F for forecasting).}
    \label{tab:ablation}
\end{table}

\subsection{Main Results}
As shown in \tableautorefname \ref{tab:main_result_classify}, iTimER consistently achieves strong performance, demonstrating its effectiveness for ISTS classification tasks. Specifically, in the \textit{imbalanced} binary classification tasks, iTimER outperforms prior methods on the P12 dataset by clear margins in both AUROC and AUPRC, indicating greater sensitivity to rare yet informative signals in sparse ICU data. It maintains competitive results on P19 and achieves SOTA performance on the balanced PAM dataset, demonstrating its consistent effectiveness across varying levels of class imbalance and missingness. Moreover, as shown in \figureautorefname~\ref{fig:efficiency}, iTimER achieves this without sacrificing efficiency. Its favorable balance between performance and complexity makes it especially suitable for ISTS.

\tableautorefname \ref{tab:main_result_interpolation} presents the performance comparison for ISTS interpolation tasks, where $30\%$ of the observation timestamps are randomly masked and the model is tasked with reconstructing them using unmasked data.  iTimER consistently achieves the best performance across all three datasets, outperforming recent SOTA methods. Its superior results, particularly on high-missingness datasets like MIMIC and PhysioNet, indicate a strong ability to capture local structure and uncertainty without relying on explicit imputation or continuous-time solvers.

\tableautorefname \ref{tab:main_result_predict} shows the performance on ISTS forecasting tasks, where we follow the common setup in \citet{zhang2024tpatchgnn} that uses the first 24 hours of data to predict the next 24 hours on PhysioNet and MIMIC datasets, and uses the first $3,000$ ms to predict the next $1,000$ ms on Human Activity data. iTimER achieves SOTA results on PhysioNet and MIMIC, and ranks among the top methods on Human Activity. These results suggest that iTimER's reconstruction-aware representation learning generalizes well to long-range extrapolation, even under significant sparsity and distribution shift.

In nearly all cases, iTimER exhibits consistently \textit{low standard deviation}, indicating it is a reliable and robust model. Its performance remains stable across varying data samples and initial conditions, highlighting its strong potential to generalize well to new, unseen data. This stability and predictability are particularly valuable in sensitive domains such as medical diagnosis, where accurate and reliable predictions are essential in clinical settings.

\subsection{Ablation Analysis and Efficiency Evaluation}

\paragraph{Ablation Analysis.} \tableautorefname \ref{tab:ablation} presents the ablation study evaluating various pseudo-observation strategies and loss components in iTimER. Here,
\textit{Baseline} performs reconstruction-based pretraining only on the observed data, without generating pseudo-observations or modeling reconstruction errors for unobserved timestamps. It serves as a lower result that lacks uncertainty modeling and signal augmentation.

We first test substitutes for $\tilde{x}_t$, i.e., the proxy for the unobserved regions: Random samples (\textit{Random}), a constant global mean (\textit{Constant}), and using only the reconstruction error (\textit{Only Error}). \textit{Random} and \textit{Constant} variants utilize uninformed or static values, leading to substantial degradation in all tasks, highlighting the importance of noise-aware generation.
While \textit{Only Error} variant relies completely on the reconstruction error component and performs better than the above two variants, but still underperforms compared to our full iTimER.
This highlights that sampling from the error distribution $\mathcal{N}\left(\mu_\epsilon, \sigma_\epsilon^2\right)$ alone introduces meaningful inductive cues, along with the fact that without anchoring to observed values, it lacks the structural grounding necessary for robust representation learning.

Subsequently, in the mixup formulation $\tilde{x}_t = \alpha_t \cdot \bar{x} + (1 - \alpha_t) \cdot \tilde{\epsilon}_t$, we test three alternatives for $\bar{x}$: setting it to zero (\textit{Zero}), the global mean (\textit{Mean}), or a local moving average (\textit{MAve}) with a window size of $5$.
Among them, the \textit{MAve} strategy achieves the best results, suggesting that preserving local dynamics benefits pseudo-observation quality. However, it introduces additional computational cost due to windowed processing and potential extra parameters. Finally, removing either the Wasserstein loss or the contrastive consistency loss leads to clear performance drops, confirming that both are essential for aligning uncertainty and ensuring representational coherence. Overall, these results validate our design choices and demonstrate that iTimER’s gains are from principled generation and multi-objective training.

\begin{figure}[t]
\centerline
{\includegraphics[width=\linewidth]{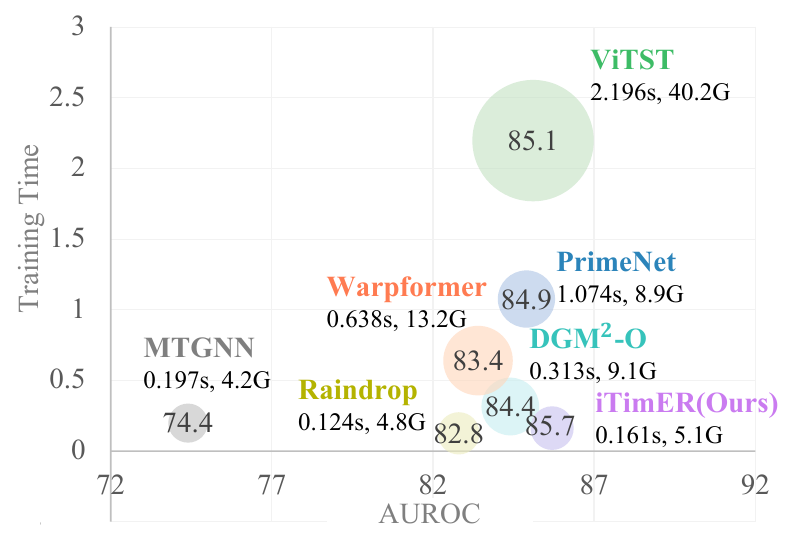}}
\caption{Efficiency comparisons in terms of Training Time (s) and Memory Usage (G) with the latest advanced models on the P12 datasets. }
\label{fig:efficiency}
\end{figure}

\paragraph{Efficiency Evaluation.} To demonstrate computational complexity, taking P12 in the classification task with a batch size of $50$ as an example, iTimER achieves a training time cost of $0.161$s per batch with $5.16$ GB of memory usage. As in \figureautorefname \ref{fig:efficiency}, the closer a circle is to the \textbf{bottom-right corner} and the \textbf{smaller} its area, the higher the model's classification accuracy, with faster training speed and lower memory usage. Our iTimER achieves lower time complexity than most other methods and uses much less memory, especially less than ViTST, which also performs well on classification tasks.

Although we achieve SOTA performance for the P12 dataset, our training time cost and memory usage are not optimal. This is primarily due to the additional overhead introduced by computing reconstruction error distributions and generating uncertainty-aware pseudo-observations during training. These steps enable iTimER to model ISTS more effectively, leading to improved performance across tasks despite the slight increase in resource consumption.

\section{Conclusion}

In this work, we propose iTimER, a novel framework for ISTS representation learning that leverages reconstruction error distributions as informative training signals. By generating pseudo-observations guided by learned error distributions and enforcing consistency across original and pseudo-observation series, iTimER effectively enriches the training signal without relying on labels or explicit imputation. Extensive experiments across three downstream tasks demonstrate its strong performance and broad applicability.

Looking forward, the learned reconstruction error distribution offers a promising plug-and-play module that can be integrated into a wide range of ISTS models. Future work will explore its utility for enhancing uncertainty modeling, guiding attention mechanisms, or improving robustness under distribution shift.

\section{Acknowledgements}
The main framework of this work was developed under the supervision of Professor Songcan Chen at Nanjing University of Aeronautics and Astronautics.

The authors wish to thank all the donors of the original datasets, and everyone who provided feedback on this work. This work is supported by the Key Program of NSFC under Grant No.62376126.

\bibliography{aaai2026}

\clearpage
\clearpage
\newpage


\end{document}